\documentclass[letterpaper]{article} 
\usepackage{aaai24}  
\usepackage{times}  
\usepackage{helvet}  
\usepackage{courier}  
\usepackage[hyphens]{url}  
\usepackage{graphicx} 
\usepackage{natbib}  
\usepackage{caption} 
\usepackage{algorithm}
\usepackage{algorithmic}
\usepackage{times}
\usepackage{epsfig}
\usepackage{graphicx}
\usepackage{amsmath}
\usepackage{amssymb}
\usepackage{float}
\usepackage[symbol]{footmisc}
\usepackage{amsmath,amsfonts,amssymb,amsthm,bm}  
\usepackage{booktabs}
\usepackage{graphicx}
\usepackage{subcaption}
\usepackage{multirow}

\usepackage{tikz}

\newcommand{\hide}[1]{} 

\newcommand\Ibb{\ensuremath{{\mathbb{I}}}}

\newcommand\Xc{\ensuremath{{\mathcal{X}}}}

\usepackage{tikz}

\usepackage{newfloat}
\usepackage{listings}
\DeclareCaptionStyle{ruled}{labelfont=normalfont,labelsep=colon,strut=off} 
\floatstyle{ruled}
\newfloat{listing}{tb}{lst}{}
\floatname{listing}{Listing}
\title{HuTuMotion: Human-Tuned Navigation of Latent Motion Diffusion Models with Minimal Feedback}
\author {
   Gaoge Han\textsuperscript{\rm 1},
    Shaoli Huang\textsuperscript{\rm 2}\footnotemark[1],
    Mingming Gong\textsuperscript{\rm 3,\rm 4}
    Jinglei Tang \textsuperscript{\rm 1}\thanks{Corresponding author: Shaoli Huang, Jinglei Tang.}
}
\affiliations {
    \textsuperscript{\rm 1}College of Information Engineering, Northwest A\&F University\\
    \textsuperscript{\rm 2}Tencent AI Lab\\
    \textsuperscript{\rm 3}School of Mathematics and Statistics, The University of Melbourne\\
    \textsuperscript{\rm 4}Mohamed bin Zayed University of Artificial Intelligence, United Arab Emirates\\
    hangaoge@nwafu.edu.cn,
    sshaol.huang@gmail.com, 
    mingming.gong@unimelb.edu.au,
    tangjinglei@nwsuaf.edu.cn 
}

\usepackage{bibentry}

\begin{document}

\maketitle

\begin{abstract}
We introduce HuTuMotion, an innovative approach for generating natural human motions that navigates latent motion diffusion models by leveraging few-shot human feedback. Unlike existing approaches that sample latent variables from a standard normal prior distribution, our method adapts the prior distribution to better suit the characteristics of the data, as indicated by human feedback, thus enhancing the quality of motion generation. Furthermore, our findings reveal that utilizing few-shot feedback can yield performance levels on par with those attained through extensive human feedback. This discovery emphasizes the potential and efficiency of incorporating few-shot human-guided optimization within latent diffusion models for personalized and style-aware human motion generation applications. The experimental results show the significantly superior performance of our method over existing state-of-the-art approaches.
\end{abstract}

\section{Introduction}
Human motion generation, a rapidly growing area of research~\cite{li2022danceformer,raab2022modi,guo2020action2motion,motionvae_ling2020character,petrovich21actor,Guo_2022_CVPR_t2m,chuan2022tm2t,Guo_2022_CVPR_humanml3d,zhang2022motiondiffuse,mdm2022human} in computer vision and artificial intelligence, has gained significant attention due to its wide-ranging applications in animation, gaming, and robotics. Recent techniques typically encompass the process of sampling latent variables $z$ from a standard normal prior distribution $p(z)$, followed by generating data $x$ using the generative probability distribution $p(x|z)$. These approaches, often based on deep generative models like Variational Autoencoders (VAEs)~\cite{chen2022mld} or Generative Adversarial Networks (GANs)~\cite{lee2019dancing}, have made significant strides in motion synthesis. However, they often fall short of capturing the real data characteristics and generating human motions that accurately reflect the input semantics. For instance, consider generating a motion from a textual input like ``an old person walking at an
average pace forward." Existing methods~\cite{chen2022mld,mdm2022human,Guo_2022_CVPR_t2m} may generate a generic ``walking at an average pace forward" motion but fail to capture the specific nuance of ``old". This limitation has hindered the advancement of more sophisticated applications, such as personalized and editable motion generation, which call for a deeper grasp of the underlying data distribution and effective incorporation of human feedback.

In light of these challenges, we propose HuTuMotion, a novel approach that seeks to improve the quality of motion generation by leveraging latent diffusion models and incorporating few-shot human feedback. The central idea of HuTuMotion is to adjust the prior distribution $p(z)$ based on human feedback rather than existing approaches of drawing from a standard normal prior distribution. This adjustment allows the latent space to capture the characteristics of the data better, thereby improving the quality and realism of the generated motions. Our method does not solely rely on an arbitrary sampling of the latent space. Instead, we carefully optimize the selection of regions in the latent space that yield more realistic and semantically meaningful human motions. To achieve this, we first strategically identify representative and diverse motion descriptions. We then utilize a unique feedback mechanism that incorporates a few-shot learning approach. In this paradigm, minimal yet effective human feedback guides the optimization process, refining the link between the descriptions and their corresponding latent distributions. Furthermore, to ensure semantic alignment between the input text and output motion, we employ a text similarity measure. During testing, this measure assesses the similarity between the input text and the representative motion descriptions. The most similar representative prior distribution is then used to sample a latent, which ultimately generates the corresponding human motion.

In addition to enhancing the general text-driven motion generation, HuTuMotion also introduces a new capability to support personalized and style-aware motion generation. This functionality allows the users to provide their specific motion style preferences, which are then incorporated into the motion generation process. Through this mechanism, our method can generate unique, individualized motions that better reflect the users' intentions and preferences, thereby opening up new possibilities for applications in areas such as interactive gaming and personalized animation. In our quantitative experiments on both the HumanML3D and KIT datasets, HuTuMotion significantly outperforms existing state-of-the-art methods. Additionally, through qualitative experiments, we observe that our method generates more natural and semantically correct motions. Importantly, when comparing the effectiveness of few-shot and extensive human feedback within our method, our results show that using few-shot human feedback achieves comparable performance to extensive feedback. This underscores the efficiency and potential of our few-shot human-guided optimization approach in the field of human motion generation applications.

The key contributions of our work are:
\begin{itemize}
\item We introduce HuTuMotion, a novel approach to improve the quality of motion generation using latent diffusion models and few-shot human feedback. To the best of our knowledge, this work is the first attempt to leverage few-shot human feedback to enhance motion generation quality.
\item HuTuMotion uniquely adjusts the prior distribution based on human feedback, optimizing the selection of regions in the latent space to yield more realistic and semantically meaningful human motions.
\item We propose a unique feedback mechanism that incorporates a few-shot learning approach and a text similarity measure to refine the link between motion descriptions and their corresponding latent distributions, enhancing semantic alignment between the input text and output motion.
\item HuTuMotion also supports personalized and style-aware motion generation, enabling users to provide specific motion style preferences that are incorporated into the motion generation process.
\end{itemize}

\section{Related Work}
{\bf Human Motion Diffusion Model.} The impressive performance of diffusion models on text-to-image tasks \cite{ho2022classifier,rombach2022high,stable_diffusion} has recently inspired the creation of diffusion-based human motion models \cite{mdm2022human,zhang2022motiondiffuse,chen2022mld,zhang2023tapmo} that are trained on the motion capture datasets using the human motion estimation methods \cite{kanazawa2018end,yu2023acr,cheng2023bopr}. MotionDiffuse~\cite{zhang2022motiondiffuse} pioneered this field as the first text-based motion diffusion model with fine-grained instructions on body parts. MDM~\cite{mdm2022human} and MLD \cite{chen2022mld} followed suit, with the former proposing a motion diffusion model on raw motion data to understand the relationship between motion and input conditions, and the latter performing the motion diffusion process in the latent space to significantly reduce computational overhead during training and inference stages. Unlike these models, our approach guides initial latent generation on a latent-based diffusion model, achieving state-of-the-art performance in text-to-motion tasks with minimal cost.

{\noindent\bf Reinforcement Learning with Human Feedback. } Reinforcement Learning with Human Feedback (RLHF) is an expanding field demonstrating substantial potential in aligning human references and model performance across language tasks \cite{stiennon2020learningsummarize,ouyang2022traininginstructgpt,chatgpt}. It typically involves using human-gathered ranking data to train a reward model, which is then used to fine-tune a Supervised Fine-Tuning (SFT) model via policy gradients. Recent applications of RLHF in the text-to-image field \cite{zhang2023hive,lee2023aligning,tang2023zeroth,xu2023imagereward,wu2023better} have shown promising text-image alignment performance. However, unlike previous works, our method uses a novel online human feedback approach, avoiding the need for large-scale human-ranked data to optimize the quality of generated human motion.

{\noindent\bf Few-Shot Learning and Generation. } Most few-shot learning methods fall into three categories: meta-learning \cite{MAML,TADAM,MatchNet}, transfer-learning \cite{yang2022few,zhang2022tip,hu2022pushing}, and feature augmentations \cite{lazarou2022tensor, Chen2018MultiLevelSF, Ye_2020_CVPR}. These methods use textual descriptions of novel classes to generate and align images, promoting the effective use of synthetic images in training few-shot learners. Recently, few-shot generation has been employed in the text-to-image task using diffusion models \cite{gal2022image, ruiz2022dreambooth,samuel2023allseedselect}. Specifically, \cite{gal2022image, ruiz2022dreambooth} learn to map a set of images to a corresponding "word" in the low-dimensional embedding space using a pre-trained model. \cite{samuel2023allseedselect} addresses long-tail learning in the presence of highly unbalanced training data by selecting optimal generation seeds from the noise space.

In our approach, we generate natural human motions by incorporating few-shot feedback and text similarity, requiring only a few-shot human feedback for searching the prior distribution of latent, thus achieving a network-free method during the inference period.

\begin{figure*}[t]
	\centering
	\includegraphics[width=0.99\linewidth]{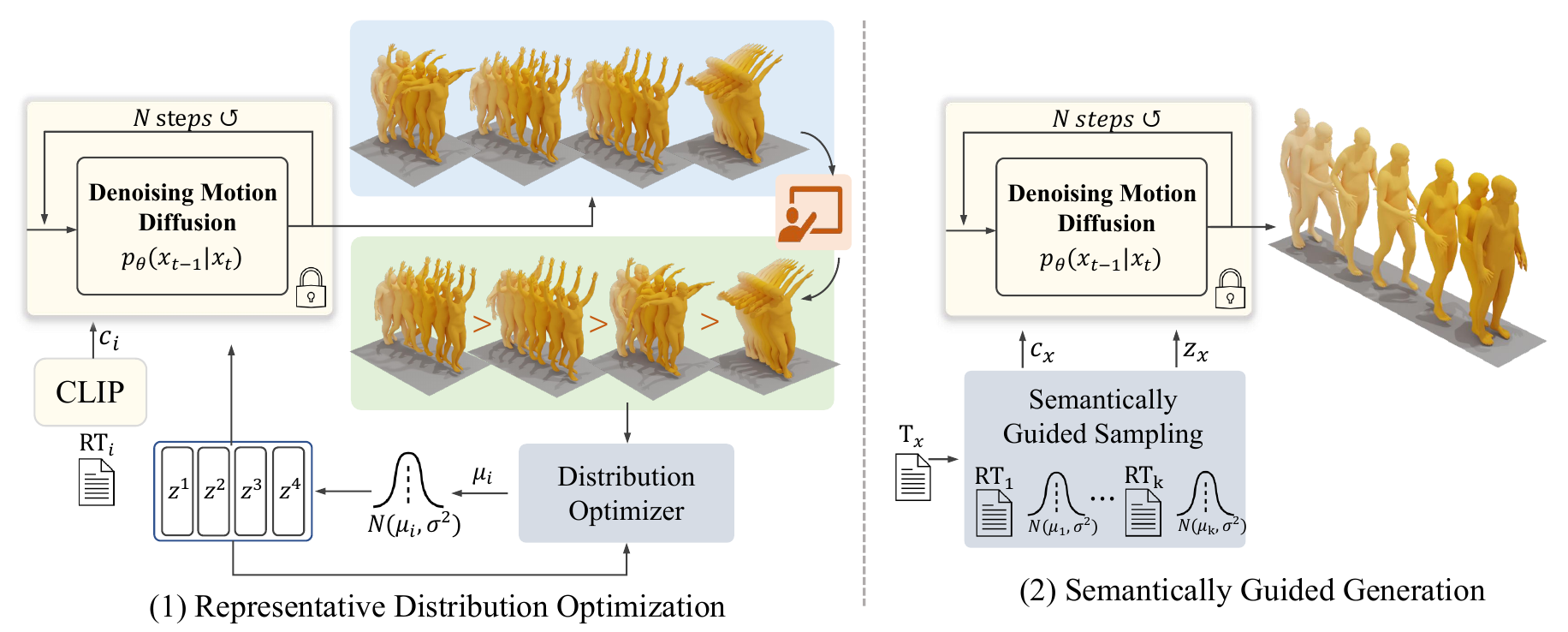}
	\caption{An overview of our framework. For simplicity, we omit the motion decoder. It consists of two stages. In representative distribution optimization, we obtain the optimized latent distribution corresponding to representative texts. In a semantically guided generation, we select one latent distribution by computing text similarity.}
	\label{fig:framework}
\end{figure*}

\section{Method}
As depicted in Figure \ref{fig:framework}, our method, HuTuMotion, is characterized by two principal steps: 1) Representative Distribution Optimization and 2) Semantic Alignment and Motion Generation. We begin with a concise introduction to the Latent Diffusion Model, followed by an in-depth explanation of these pivotal steps.

\subsection{Overview of Latent Diffusion Models} Latent diffusion models (LDMs) has achieved great success in text-to-motion task, such as Stable Diffusion \cite{stable_diffusion}. Different from the diffusion model, LDMs perform diffusion process in latent space using an extra denoising U-Net. Low-dimensional space is better suited for likelihood-based generative models, as it allows them to concentrate on the crucial semantic representation of the data and trains in a lower-dimensional space that is computationally more efficient. MLD \cite{chen2022mld} is the first latent-based motion diffusion model for text-to-motion synthesis. MLD design a transformer-based conditional denoiser $\epsilon_\theta$. Its conditional objective can be expressed as: 

\begin{equation}
    \mathcal{L}_{\textbf{MLD}}:=\mathbb{E}_{\epsilon, t, c}\left[\left\|\epsilon-\epsilon_\theta\left(z_t, t, \tau_\theta(c)\right)\right\|_2^2\right],
\end{equation}

where $\epsilon \sim \mathcal{N}(0, I)$, $z_t$ is the latent in time step $t$ and $ \tau_\theta(c)$ denotes the CLIP \cite{radford2021learningCLIP} text encoder with embedded conditions $c$.  MLD uses the DDIM \cite{song2020denoisingddim} as the sampler. It is worth noting that the denoising diffusion process becomes deterministic and solely on the latent embedding for ODE-based diffusion samplers such as DDIM \cite{song2020denoisingddim}, DPM-solver \cite{lu2022dpm} and DPM-solver++ \cite{lu2022dpm++}. We optimize the latent distribution based on MLD.

\subsection{Representative  Distribution Optimization}
\label{humantuning}
Our methodology starts by identifying a set of representative and diverse motion descriptions. We then incorporate few-shot human feedback to adjust the prior distribution through a two-stage optimization process.

{\bf Representative Motion Descriptions.} Representative motion descriptions can be defined as descriptors that capture the wide range and diversity of human motions. A straightforward approach for selecting these descriptions involves employing a K-means clustering algorithm on the training dataset. In this process, the number of clusters denoted as $K$, is predetermined based on the diversity present in the data. Each cluster's centroid is then considered a representative motion description.
However, our experimental observations indicate that a representative motion description does not necessarily need to come from a specific training dataset. Asking a large language model (ChatGPT \cite{chatgpt}) to generate representative texts can also yield comparable results. The detailed process can be found in the supplementary material.

{\bf Latent Optimization via Human Feedback.}
The central innovation of HuTuMotion lies in the optimization of the latent variables from the prior distribution. Unlike conventional methods that draw samples from a standard normal prior distribution, we adjust the latent input from the prior distribution based on few-shot human feedback in the form of rankings. 

Our goal is to ascertain an optimal value of $z$ that yields the minimum score for the function $f(z,c,t)$, wherein $c$ represents the text embedding and $t$ signifies the diffusion steps. Given that $c$ and $t$ remain constant throughout the optimization process, we will omit them in the ensuing discussions. Here, the score of $f(z)$ should reflect the quality of the generated motion from the input $z$ according to human judgment. The lower the score, the better the generated motion. The specific form of $f(z)$ is not predefined and is implicitly determined by human feedback.  To formalize this, we can express it as an optimization problem: $\min\limits_{z\in \mathbb{R}^d} f(z).$

Given that $f(z)$ functions as a black box, we do not have direct access to its internal workings. Instead, we can interface with it through a ranking oracle. This oracle provides insight into the function by sorting the scores of generated outcomes, which can be thought of as motions in a particular context.

Due to the constraints imposed by the ranking oracle, the optimization problem at hand can be recast as an $(m,k)$-ranking oracle optimization problem. This type of problem involves determining an optimal ranking of a set of items constrained by the limited information provided by the oracle. Therefore, the challenge lies not only in optimizing $f(z)$ but also in intelligently querying the oracle to explore the search space efficiently.

To solve this problem, we adopt the zeroth-order optimization algorithm~\cite{tang2023zeroth} that obtains the descent direction using a rank-based random estimator. The estimator converts the $(m,k)$-ranking oracles' input and output into a directed acyclic graph (DAG), $\mathcal{G}=(\mathcal{N},\mathcal{E})$, where $\mathcal{N} = \{1,...,m\}$ and $\mathcal{E} = {(i,j)~|~f(z_i) < f(z_j)}$. With access to an $(m,k)$-ranking oracle $O(m,k)$ and a starting point $z$, we query $O(m,k)$ using $x_i = x + \mu \xi_i$, where $\xi_i \sim \mathcal{N}(0, I)$ for $i = 1,...,m$. The rank-based gradient estimator, constructed using the ranking information from $O(m,k)$, is:
\begin{equation}
\tilde{g}(z) = \frac{1}{|\mathcal{E}|}\sum\limits_{(i,j)\in \mathcal{E}}\frac{\xi_j-\xi_i}{\mu}.
\label{eq:gradient}
\end{equation}

Upon the completion of the initial optimization process, we continue to refine the optimal latent variable $z$. The comprehensive steps for this procedure are outlined in Algorithm~\ref{alg:two-stage}.

{\bf Optimal Representative Prior Distribution.} We have noted a significant correlation in our observations: texts that are closely aligned in the embedding space also demonstrate similar proximity in their optimal latents. Detailed results supporting this observation can be found in the accompanying supplementary material. Based on this, we propose a method to construct an optimal prior distribution for texts that are near a given representative text ($RT$). We use the optimal latent of a representative text as the mean and specify a relatively small standard deviation to construct a Gaussian distribution. This Gaussian distribution then serves as the optimal prior distribution for texts that are close to the representative text in the embedding space.

\subsection{Semantic Alignment and Motion Generation}
\label{inference}

By employing representative distribution optimization coupled with human ranking information, we acquire pair sets consisting of representative texts and their corresponding latent distributions. Given an input text, we initially calculate its cosine similarity with the representative texts. Subsequently, we select the latent distribution that corresponds to the representative text closest in similarity. This approach serves as an efficient strategy to enhance overall motion quality, as it does not necessitate significant resources to fine-tune the model.

{\bf Semantically Guided Sampling.} Contrary to the MLD approach which samples latent from $\mathcal{N}(0, I)$, we instead sample latent from $\mathcal{N}(z^{**}_m,\sigma)$, where $z^{**}_m$ is the optimal latent determined for the representative text $RT_m$. When given an input text, we compute the index $m$ by measuring the cosine similarity between  $c_x$ and $\{{c_1, ..., c_k}\}$ corresponding to the input text $T_x$ and the set of representative texts $\{RT_1, ..., RT_k\}$, respectively. In this context, $c_x$ and $\{{c_1, ..., c_k}\}$ denote the text embeddings obtained from the CLIP text encoder. The maximum index $m$ is then computed using the following equation:

\begin{equation}
m = \arg \max_i \frac{c_x \cdot c_i}{||c_x|| \cdot ||c_i||}. \quad \text{for } i \in {1, ..., k}
\end{equation}

This allows us to select the representative text that aligns most closely with the given input text, ensuring a more effective and relevant sampling of the latent variable.

\begin{algorithm}[H]
    \small
        \begin{algorithmic}[1]
          \REQUIRE Objective function $f$ (Evaluated by human),  number of queries $m$, stepsize $\eta$, smoothing parameter $\mu_1$, $\mu_2$, $\mu_3$, shrinking rate $\gamma\in(0,1)$.

          \STATE Initialize the reference point $z^*$ with all-zero vectors. 
          \STATE Initialize the gradient memory $\bar g$ with all-zero vectors. 
          \STATE Set $\tau=0$.
          \STATE Choose one sample from the representative texts provided by ChatGPT.
          \STATE Sample $m$ i.i.d. starting point input $ \Xc_1= \{\xi_{1},\cdots,\xi_{m}\}$  from $\mathcal{N}(0, \mu_1 I)$.
        \WHILE{not select the best motion by human}

        \STATE Query  $O_f^{(m,k)}$ with input $\Xc_1$ for $2\leq k\leq m$. Denote $\Ibb_1$ as the output.
        \STATE Set $z^{*}$ to be the weighted $\Xc_1$ using the ranking information $\Ibb_1$.
        \STATE Compute the gradient $\tilde{g}$ using the ranking information $\Ibb_1$ according to the equation~\ref{eq:gradient}.
        \STATE $\bar g=(\tau\bar g +\tilde{g} )/(\tau+1)$
        \STATE $\tau=\tau+1$
        \STATE Sample $m$ i.i.d. direction $ \{\psi_{1},\cdots,\psi_{m}\}$  from $\mathcal{N}(0, \mu_2 I)$.
        \STATE $\Xc_1=\{z^*-\eta \bar g + \psi_{1},z^*-\eta\gamma \bar  g +  \psi_{2},...,z^*-\eta\gamma^{m-1} \bar  g  + \psi_{m}\}$

        \ENDWHILE
        \STATE Set $z^{**}$ to be the best point in $\Xc_1$ with minimal objective value using the ranking information $\Ibb_1$.
        \WHILE{not exit by human}
        
        \STATE Sample $m$ i.i.d. direction $ \{\psi_{1},\cdots,\psi_{m}\}$  from $\mathcal{N}(0, \mu_3 I)$.
        \STATE Query $O_f^{(m,1)}$ with input $\Xc_2=\{z^{**} + \psi_{1},z^{**} +  \psi_{2},...,z^{**} + \psi_{m}\}$. Denote $\Ibb_2$ as the output.
        \STATE Set $z^{**}$ to be the best point in $\Xc_2$ with minimal objective value using the ranking information $\Ibb_2$.
        \ENDWHILE

        \end{algorithmic}
        \caption{Distribution optimization for representative texts}
        \label{alg:two-stage}
        \end{algorithm}

{\bf Inference (Motion Generation).} By sampling from the optimal distribution $\mathcal{N}(z^{**}_m,\sigma^2)$, we obtain $z_x$. Subsequently, we input $z_x$ and $c_x$ into the DDIM sampler \cite{song2020denoisingddim} to facilitate the denoising motion diffusion process. It's important to note that the standard deviation $\sigma$ acts as our hyper-parameter. We further examine and discuss its impact in the Ablation Studies section.

\begin{table*}[t]
\resizebox{\linewidth}{!}{%
\begin{tabular}{@{}lccccccc@{}}
\toprule
\multirow{2}{*}{Methods} & \multicolumn{3}{c}{R Precision $\uparrow$}                                                                                                                & \multicolumn{1}{c}{\multirow{2}{*}{FID$\downarrow$}} & \multirow{2}{*}{MM Dist$\downarrow$}              & \multirow{2}{*}{Diversity$\uparrow$}           & \multirow{2}{*}{MModality}       \\ \cmidrule(lr){2-4}
              & \multicolumn{1}{c}{Top 1} & \multicolumn{1}{c}{Top 2} & \multicolumn{1}{c}{Top 3} & \multicolumn{1}{c}{}                     &                          &                            &                            \\ \midrule
Real &
  $0.511^{\pm.003}$ &
  $0.703^{\pm.003}$ &
  $0.797^{\pm.002}$ &
  $0.002^{\pm.000}$ &
  $2.974^{\pm.008}$ &
  $9.503^{\pm.065}$ &
  \multicolumn{1}{c}{-}
  \\ \midrule
Seq2Seq \cite{plappert2018learning} &
  $0.180^{\pm.002}$ &
  $0.300^{\pm.002}$ &
  $0.396^{\pm.002}$ &
  $11.75^{\pm.035}$ &
  $5.529^{\pm.007}$ &
  $6.223^{\pm.061}$ &
  \multicolumn{1}{c}{-} \\
LJ2P \cite{ahuja2019language2pose}&
  $0.246^{\pm.001}$ &
  $0.387^{\pm.002}$ &
  $0.486^{\pm.002}$ &
  $11.02^{\pm.046}$ &
  $5.296^{\pm.008}$ &
  $7.676^{\pm.058}$ &
  \multicolumn{1}{c}{-} \\
T2G\cite{bhattacharya2021text2gestures} &
  $0.165^{\pm.001}$ &
  $0.267^{\pm.002}$ &
  $0.345^{\pm.002}$ &
  $7.664^{\pm.030}$ &
  $6.030^{\pm.008}$ &
  $6.409^{\pm.071}$ &
  \multicolumn{1}{c}{-} \\
Hier \cite{ghosh2021synthesis}&
    $0.301^{\pm.002}$ &
    $0.425^{\pm.002}$ &
    $0.552^{\pm.004}$ &
    $6.532^{\pm.024}$ &
    $5.012^{\pm.018}$ &
    $8.332^{\pm.042}$ &
    \multicolumn{1}{c}{-} \\
TEMOS \cite{petrovich22temos}&
  $0.424^{\pm.002}$ &
  $0.612^{\pm.002}$ &
  $0.722^{\pm.002}$ &
  $3.734^{\pm.028}$ &
  $3.703^{\pm.008}$ &
  $8.973^{\pm.071}$ &
  $0.368^{\pm.018}$ \\

T2M \cite{Guo_2022_CVPR_t2m}&
  $0.457^{\pm.002}$ &
  $0.639^{\pm.003}$ &
  $0.740^{\pm.003}$ &
  $1.067^{\pm.002}$ &
  $3.340^{\pm.008}$ &
  $9.188^{\pm.002}$ &
  $2.090^{\pm.083}$ \\
MDM  \cite{mdm2022human}&
  $0.320^{\pm.005}$ &
  $0.498^{\pm.004}$ &
  $0.611^{\pm.007}$ &
  $0.544^{\pm.044}$ &
  $5.566^{\pm.027}$ &
  $9.559^{\pm.086}$ &
  $2.799^{\pm.072}$ \\

MLD \cite{chen2022mld} &
  $0.481^{\pm.003}$ &
  $0.673^{\pm.003}$ &
  $0.772^{\pm.002}$ &
  $0.473^{\pm.013}$ &
  $3.196^{\pm.010}$ &
  $9.724^{\pm.082}$ &
  $2.413^{\pm.079}$ \\
 \midrule
 
$\rm Ours^\ast $ &
  $\boldsymbol{0.500}^{\pm.003}$ &
  $\underline{0.686}^{\pm.002}$ &
  $\underline{0.782}^{\pm.002}$ &
  $\boldsymbol{0.224}^{\pm.006}$ &
  $\boldsymbol{3.058}^{\pm.009}$ &
  $\underline{9.745}^{\pm.073}$ &
  $0.966^{\pm.046}$\\

Ours &
  $\underline{0.497}^{\pm.002}$ &
  $\boldsymbol{0.689}^{\pm.002}$ &
  $\boldsymbol{0.785}^{\pm.002}$ &
  $\underline{0.295}^{\pm.006}$ &
  $\underline{3.093}^{\pm.007}$ &
  $\boldsymbol{9.828}^{\pm.091}$ &
  $1.019^{\pm.054}$

   \\ \bottomrule
\end{tabular}%
}

\caption{Comparison of text-to-motion synthesis on HumanML3D~\cite{Guo_2022_CVPR_humanml3d} dataset. $\ast$ means using the texts of cluster’s centroid of K-means. These metrics are evaluated by the motion encoder from \cite{Guo_2022_CVPR_t2m}. For each metric, we repeat the evaluation 20 times and report the average with a 95\% confidence interval. We employ real motion as a reference and sort all approaches by descending FIDs. \textbf{Bold} and \underline{underline} indicate the best and the second best result.}

\label{tab:tm:comp:humanml3d}
\end{table*}

\begin{table*}[t]
\resizebox{\linewidth}{!}{%
\begin{tabular}{@{}lcccccccc@{}}
\toprule
\multirow{2}{*}{Methods} & \multicolumn{3}{c}{R Precision $\uparrow$}                                                                                                                & \multicolumn{1}{c}{\multirow{2}{*}{FID$\downarrow$}} & \multirow{2}{*}{MM Dist$\downarrow$}              & \multirow{2}{*}{Diversity$\uparrow$}           & \multirow{2}{*}{MModality}              \\ \cmidrule(lr){2-4}
              & \multicolumn{1}{c}{Top 1} & \multicolumn{1}{c}{Top 2} & \multicolumn{1}{c}{Top 3} & \multicolumn{1}{c}{}                     &                          &                            &                            \\ \midrule
Real &
  $0.424^{\pm.005}$ &
  $0.649^{\pm.006}$ &
  $0.779^{\pm.006}$ &
  $0.031^{\pm.004}$ &
  $2.788^{\pm.012}$ &
  $11.08^{\pm.097}$ &
  \multicolumn{1}{c}{-}
  \\ \midrule
Seq2Seq\cite{plappert2018learning} &
  $0.103^{\pm.003}$ &
  $0.178^{\pm.005}$ &
  $0.241^{\pm.006}$ &
  $24.86^{\pm.348}$ &
  $7.960^{\pm.031}$ &
  $6.744^{\pm.106}$ &
  \multicolumn{1}{c}{-} \\
T2G\cite{bhattacharya2021text2gestures} &
  $0.156^{\pm.004}$ &
  $0.255^{\pm.004}$ &
  $0.338^{\pm.005}$ &
  $12.12^{\pm.183}$ &
  $6.964^{\pm.029}$ &
  $9.334^{\pm.079}$ &
  \multicolumn{1}{c}{-} \\
LJ2P \cite{ahuja2019language2pose}&
  $0.221^{\pm.005}$ &
  $0.373^{\pm.004}$ &
  $0.483^{\pm.005}$ &
  $6.545^{\pm.072}$ &
  $5.147^{\pm.030}$ &
  $9.073^{\pm.100}$ &
  \multicolumn{1}{c}{-} \\
Hier \cite{ghosh2021synthesis}&
  $0.255^{\pm.006}$ &
  $0.432^{\pm.007}$ &
  $0.531^{\pm.007}$ &
  $5.203^{\pm.107}$ &
  $4.986^{\pm.027}$ &
  $9.563^{\pm.072}$ &
  $2.090^{\pm.083}$ \\
TEMOS \cite{petrovich22temos}&
    $0.353^{\pm.006}$ & 
    $0.561^{\pm.007}$ & 
    $0.687^{\pm.005}$ & 
    $3.717^{\pm.051}$ & 
    $3.417^{\pm.019}$ & 
    $10.84^{\pm.100}$ & 
    $0.532^{\pm.034}$ \\
T2M \cite{Guo_2022_CVPR_t2m}&
  $0.370^{\pm.005}$ &
  $0.569^{\pm.007}$ &
  $0.693^{\pm.007}$ &
  $2.770^{\pm.109}$ &
  $3.401^{\pm.008}$ &
  ${10.91}^{\pm.119}$ &
  $1.482^{\pm.065}$ \\
MDM \cite{mdm2022human}&
  $0.164^{\pm.004}$ &
  $0.291^{\pm.004}$ &
  $0.396^{\pm.004}$ &
  ${0.497}^{\pm.021}$ &
  $9.191^{\pm.022}$ &
  $10.85^{\pm.109}$ &
  $1.907^{\pm.214}$ \\

MLD \cite{chen2022mld} &
${0.390}^{\pm.008}$ & 
${0.609}^{\pm.008}$ & 
${0.734}^{\pm.007}$ & 
${0.404}^{\pm.027}$ & 
${3.204}^{\pm.027}$ & 
$10.80^{\pm.117}$ & 
$2.192^{\pm.071}$\\
\midrule

$\rm Ours^\ast$ &
  $\underline{0.409}^{\pm.004}$ &
  $\boldsymbol{0.640}^{\pm.004}$ &
  $\underline{0.766}^{\pm.005}$ &
  $\boldsymbol{0.201}^{\pm.064}$ &
  $\boldsymbol{3.082}^{\pm.025}$ &
  $\underline{10.88}^{\pm.086}$ &
  $0.901^{\pm.035}$\\

Ours  &
$\boldsymbol{0.411}^{\pm.005}$ & 
$\underline{0.638}^{\pm.006}$ & 
$\boldsymbol{0.768}^{\pm.006}$ & 
$\underline{0.224}^{\pm.045}$ & 
$\underline{3.098}^{\pm.025}$ & 
$\boldsymbol{10.96}^{\pm.090}$ & 
${0.914}^{\pm.039}$
   \\ \bottomrule
\end{tabular}%
}

\caption{Comparison of text-to-motion synthesis on KIT~\cite{Plappert2016kit} dataset. $\ast$ means using the texts of cluster’s centroid of K-means. Reported metrics are the same as Table \ref{tab:tm:comp:humanml3d}. \textbf{Bold} and \underline{underline} indicate the best and the second best result.}

\label{tab:tm:comp:kit}
\end{table*}

\subsection{Expanding the Scope: Personalized and Style-Aware Generation} Our method is primarily designed to enhance the generation quality of general human motions. However, its versatility allows for straightforward extensions to accommodate new tasks. These include but are not limited to, personalized and style-aware motion generation, further demonstrating the adaptability and potential of our approach.

{\bf Personalized Motion Generation.}
Personalized Motion Generation is a task in human motion generation that focuses on generalizing motions that align with user preferences. In this context, suppose we have an output set $\{(RT_1,z^{**}_1), ..., (RT_k,z^{**}_k)\}$ derived from our method and a user-provided text set $T^1_U, ..., T^l_U$ that reflects a specific preference for a desired motion style. To implement Personalized Motion Generation, we begin by identifying the closest $RT_i$ for each text $T^j_U$ and use $z^{**}_i$ as the starting point in Algorithm~\ref{alg:two-stage} to conduct feedback optimization. The final optimized result $z^{**U}_i$ will then replace the original $z^{**}_i$. By processing all the user sets in this manner, we generate the final output set $\{(RT_1,z^{**U}_1), ..., (RT_k,z^{**U}_k)\}$ which can be used to generate motions that align with the user's preferences.

{\bf Style-Aware Motion Generation.}
Style-aware Motion Generation pertains to the task of producing motions that embody a specified style, given input text containing style descriptions. Different from personalization, the stylized prompt presents a long-tail distribution in the HumanML3D and KIT datasets. We observe that using the stylized prompt as input, the MLD model fails to generate motion consistent with the stylistic semantics, as shown in Fig. \ref{fig:application}. This differs from Personalized Motion Generation in that it need to feedback the text with style Words. Initially, it identifies diverse texts corresponding to the same style descriptor $ST_i$. Following this, it carries out optimization from scratch to secure the optimal latent. In this context, we can amass a variety of styles and establish a comprehensive set $\{(ST_1,z^{**}_1), ..., (ST_M,z^{**}_M)\}$ by employing the previously mentioned optimization process. This approach allows us to generate a broad spectrum of stylized motions.

\section{Experiments}
In this section, we provide extensive experimental results. Firstly, we introduce the datasets, implementation details, and evaluation metrics. Secondly, we show the qualitative and quantitative results compared with the state-of-the-art approaches. Finally, we perform ablation studies. More qualitative results are provided in the supplementary material.

\subsection{Datasets and Evaluation Metrics}
We experiment with two text-to-motion synthesis datasets: HumanML3D~\cite{Guo_2022_CVPR_humanml3d} and KIT~\cite{Plappert2016kit}. 

\textbf{KIT} comprises 3,911 motion sequences and 6,278 text annotations, with each motion linked to one to four descriptions. The dataset downsampled to 12.5 FPS, is partitioned into 80\% training, 5\% validation, and 15\% test sets.

\textbf{HumanML3D} is the largest 3D human motion-language dataset, containing 14,616 human motions and 44,970 text descriptions. Each motion pairs with at least three descriptions. Motions, re-scaled to 20 FPS and spanning between 2 and 10 seconds, are divided similarly to KIT.

We evaluate text-to-motion models based on metrics as described in \cite{Guo_2022_CVPR_humanml3d}: \textbf{R Precision}: This measures the accuracy of the top-1, top-2, and top-3 ranked Euclidean distances between one motion sequence and 32 text descriptions. \textbf{Frechet Inception Distance (FID):} This assesses the feature distribution distance between generated and real motions using a feature extractor. \textbf{Multimodal Distance (MM-Dist):} This metric calculates the average Euclidean distance between the generated motion feature and each text feature. \textbf{Diversity:} We measure the diversity within a motion set by calculating the average Euclidean distance between features of randomly selected motion pairs. \textbf{Multimodality (MModality):} This evaluates the diversity of motion generated for one text description by averaging the Euclidean distances between features of generated motion pairs.

\subsection{Implementation Details} 
We implement our method using the state-of-the-art latent human diffusion model, MLD \cite{chen2022mld}. Our representative distribution optimization and semantically guided generation are conducted on a single NVIDIA GeForce RTX 2080 Ti GPU, with text embedding and latent dimensions set to 768 and 256, respectively. We set $\sigma$ to 0.2 for latent sampling and use DDIM \cite{song2020denoisingddim} as the denoising motion diffusion sampler. All other settings are consistent with MLD \cite{chen2022mld}. We obtain five representative textual descriptions using ChatGPT \cite{chatgpt} (refer to supplementary material) and optimize the representative distribution with human feedback according to  Algorithm \ref{alg:two-stage}. The experimental details considering Algorithm \ref{alg:two-stage} are provided in the supplementary material.

\begin{figure}[t]
	\centering
	\includegraphics[width=0.99\linewidth]{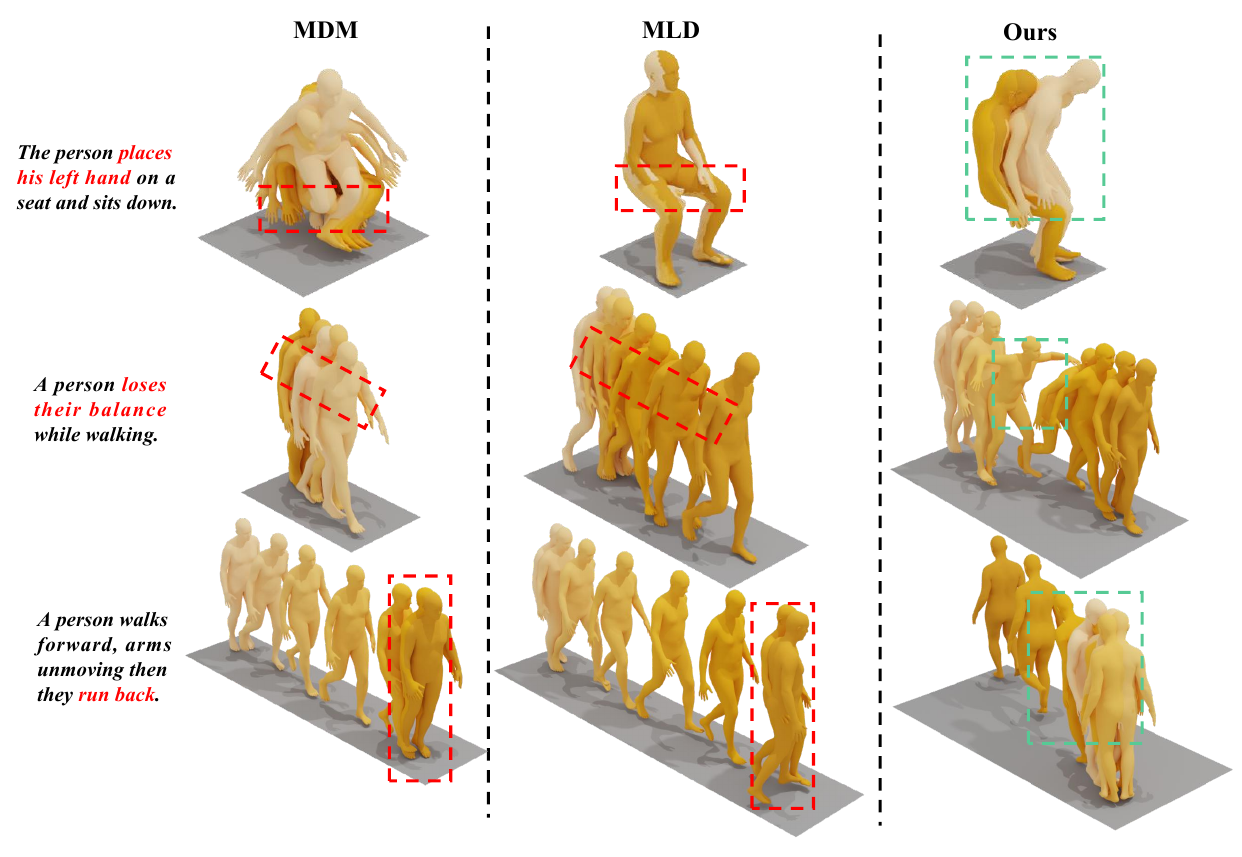}
	\caption{Qualitative results on HumanML3D~\cite{Guo_2022_CVPR_humanml3d} dataset. The darker colors indicate the later frame in time.}
	\label{fig:vis}
\end{figure}

\subsection{Comparison with State-of-the-art Methods}
In this section, we present the quantitative and qualitative results compare to existing state-of-the-art methods \cite{plappert2018learning,ahuja2019language2pose,bhattacharya2021text2gestures,ghosh2021synthesis,petrovich22temos,Guo_2022_CVPR_t2m,mdm2022human,chen2022mld} on the test set of HumanML3D \cite{Guo_2022_CVPR_humanml3d} and KIT \cite{Plappert2016kit}. Our method is implemented based on MLD \cite{chen2022mld}.

{\bf Quantitative Results Comparison.} 
The comparison results presented in Table \ref{tab:tm:comp:kit} and Table \ref{tab:tm:comp:humanml3d} on the HumanML3D \cite{Guo_2022_CVPR_humanml3d} and KIT \cite{Plappert2016kit} test sets illustrate the superior performance of our approach. It significantly outperforms other state-of-the-art methods, achieving the best scores in R Precision, FID, MM Dist, and Diversity metrics. This performance consistency across both datasets underlines the robustness of our proposed method. Although we didn't observe a consistent improvement in MModality, a higher MModality doesn't necessarily denote superior algorithm performance, as it can lead to semantically incorrect motions. Also, our MModality metric is significantly influenced by the hyperparameter $\sigma$, which we explore further in Ablation Studies section.

\begin{figure}[t]
	\centering
	\includegraphics[width=0.99\linewidth]{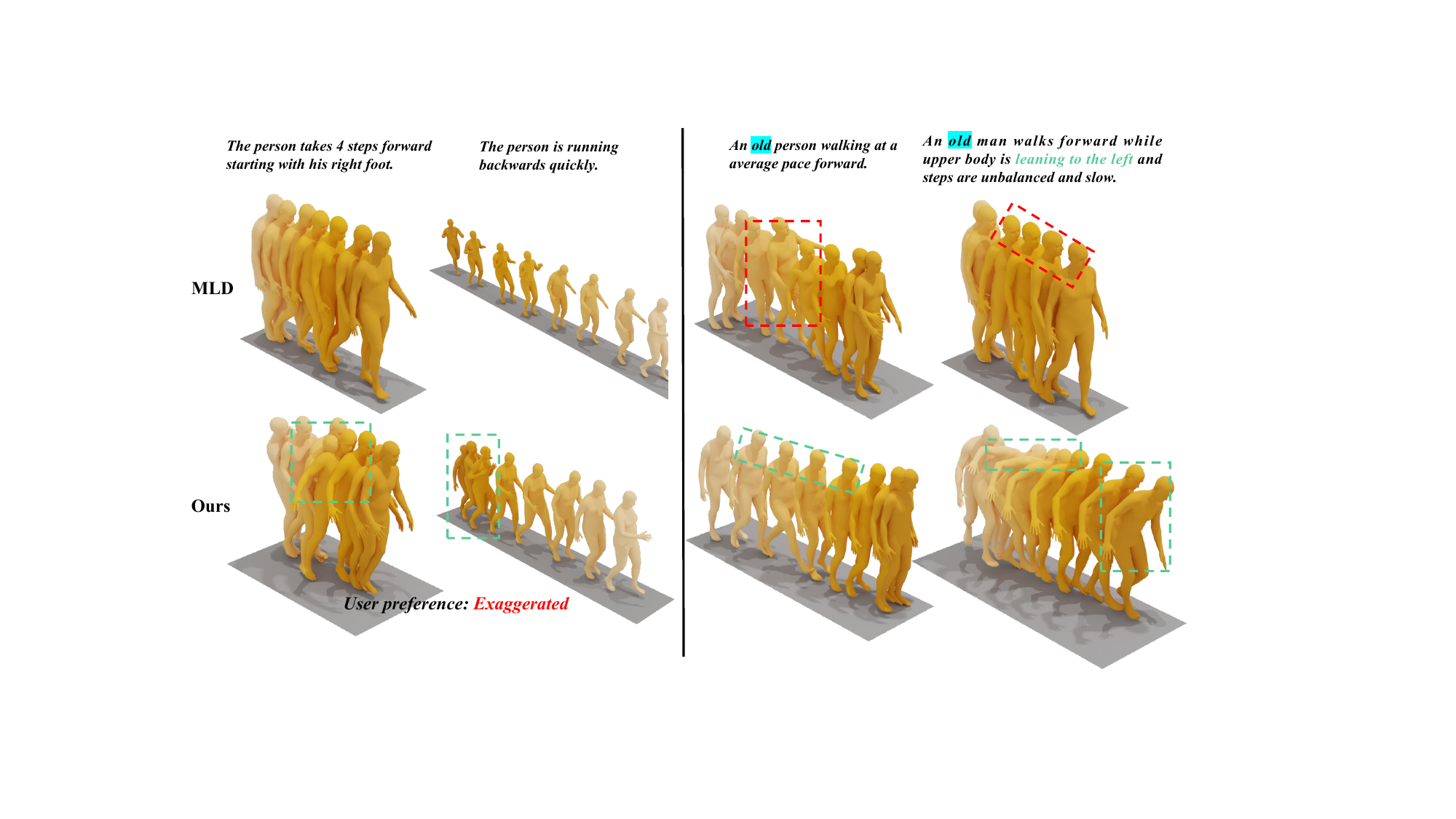}
	\caption{Personalized (left) and style-aware motion generation (right). More results are provided in the supplementary material. The darker colors indicate the later frame in time.}
	\label{fig:application}
\end{figure}

\begin{table*}[t]
\resizebox{\linewidth}{!}{%
\begin{tabular}{@{}lccccccc@{}}
\toprule
\multirow{2}{*}{Methods} & \multicolumn{3}{c}{R Precision $\uparrow$}                                                                                                                & \multicolumn{1}{c}{\multirow{2}{*}{FID$\downarrow$}} & \multirow{2}{*}{MM Dist$\downarrow$}              & \multirow{2}{*}{Diversity$\uparrow$}           & \multirow{2}{*}{MModality}              \\ \cmidrule(lr){2-4}
              & \multicolumn{1}{c}{Top 1} & \multicolumn{1}{c}{Top 2} & \multicolumn{1}{c}{Top 3} & \multicolumn{1}{c}{}                     &                          &                            &                            \\ \midrule
Real &
  $0.511^{\pm.003}$ &
  $0.703^{\pm.003}$ &
  $0.797^{\pm.002}$ &
  $0.002^{\pm.000}$ &
  $2.974^{\pm.008}$ &
  $9.503^{\pm.065}$ &
  \multicolumn{1}{c}{-}
  \\ \midrule

Ours ($\sigma = 0.1$) &
  $\underline{0.496}^{\pm.002}$ &
  ${0.687}^{\pm.002}$ &
  $\underline{0.784}^{\pm.001}$ &
  $\boldsymbol{0.258}^{\pm.005}$ &
  $\boldsymbol{3.091}^{\pm.006}$ &
  ${9.780}^{\pm.089}$ &
  $0.637^{\pm.038}$\\

Ours ($\sigma = 0.2$) &
  $\boldsymbol{0.497}^{\pm.002}$ &
  $\boldsymbol{0.689}^{\pm.002}$ &
  $\boldsymbol{0.785}^{\pm.002}$ &
  $\underline{0.295}^{\pm.006}$ &
  $\underline{3.093}^{\pm.007}$ &
  ${9.828}^{\pm.091}$ &
  $1.019^{\pm.054}$\\

Ours ($\sigma = 0.3$) &
  ${0.494}^{\pm.001}$ &
  $\underline{0.688}^{\pm.001}$ &
  $\boldsymbol{0.785}^{\pm.002}$ &
  ${0.356}^{\pm.007}$ &
  ${3.103}^{\pm.006}$ &
  $\underline{9.867}^{\pm.092}$ &
  $1.276^{\pm.062}$\\

 Ours ($\sigma = 0.4$) &
  ${0.491}^{\pm.002}$ &
  ${0.685}^{\pm.001}$ &
  ${0.781}^{\pm.002}$ &
  ${0.437}^{\pm.006}$ &
  ${3.130}^{\pm.007}$ &
  $\boldsymbol{9.882}^{\pm.087}$ &
  $1.480^{\pm.070}$\\

 Ours ($\sigma = 0.5$) &
  ${0.481}^{\pm.002}$ &
  ${0.674}^{\pm.001}$ &
  ${0.772}^{\pm.001}$ &
  ${0.554}^{\pm.010}$ &
  ${3.190}^{\pm.009}$ &
  ${9.822}^{\pm.085}$ &
  $1.680^{\pm.076}$

   \\ \bottomrule
\end{tabular}%
}

\caption{Effect of $\sigma$ on HumanML3D~\cite{Guo_2022_CVPR_humanml3d} dataset. Reported metrics are the same as Table \ref{tab:tm:comp:humanml3d}. \textbf{Bold} and \underline{underline} indicate the best and the second best result.}

\label{tab:abla:sigma}
\end{table*}

\begin{figure*}[t]
	\centering
	\includegraphics[width=0.9\linewidth]{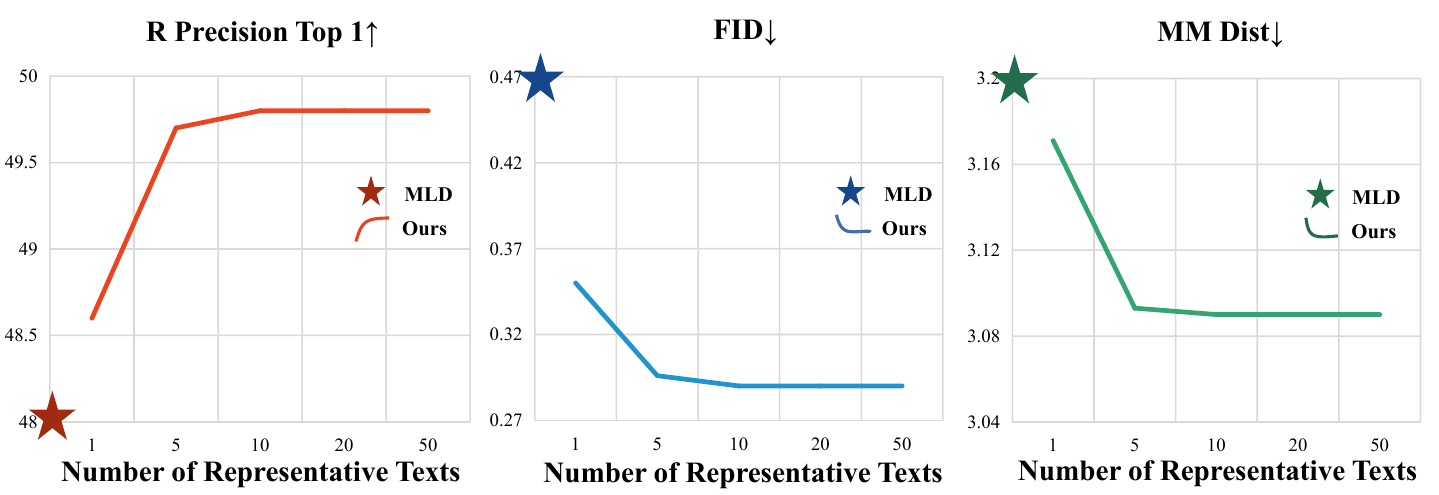}
	\caption{Effect of varying the number of representative texts  on R Precision Top 1, FID and MM Dist.}
	\label{fig:number}
\end{figure*}

{\bf Qualitative Results Comparison.}
Figure \ref{fig:vis} displays qualitative results on the HumanML3D dataset \cite{Guo_2022_CVPR_humanml3d}. The examples clearly demonstrate that our method generates more natural and semantically accurate motions compared to the lower-quality motions produced by MLD and MDM. For instance, given the text ``a person loses their balance while walking." MLD and MDM fail to capture the ``losing balance" action. Similarly, with the input, ``the person places his left hand on a seat and sits down." MLD generates a static motion while MDM exhibits distorted movements. And for the phrase ``a person walks forward, arms unmoving, then they run back." MLD and MDM cannot produce the ``run back" motion sequence.

{\bf Personalized and Style-aware Generation.} 
Figure \ref{fig:application} showcases the personalized and style-aware motion generation capabilities of our method. In personalization, given the same text input, MLD produces standard motions, whereas our method can generate exaggerated actions in line with human preferences. Regarding text-guided stylization, MLD fails to generate style-aware motion. For instance, given the input ``an old man walks forward while the upper body is leaning to the left and steps are unbalanced and slow", MLD fails to generate a sequence exhibiting the desired ``leaning to the left" and ``old" characteristics. In contrast, our method can generate semantically consistent motions while also reflecting an elderly style.

\subsection{Ablation Studies}
\label{ablation}
In this section, we first examine the influence of the number of human feedback samples. Following that, we investigate the optimal hyper-parameters for $\sigma$.

{\bf Effect of The Number of Representative Texts.}
Fig. \ref{fig:number} examines the impact of varying the number of representative texts on semantically guided generation using the HumanML3D dataset. The metrics of R Precision Top 1, FID, and MM Dist are reported for a range of 1 to 50 representative texts. We found that increasing the text count does not guarantee performance enhancement. To balance performance and distribution optimization, we opted for 5 representative texts as our default implementation setting.

{\bf Effect of The Standard Deviation $\sigma$.}
Table \ref{tab:abla:sigma} reveals the effects of varying $\sigma$ between 0.1 and 0.5. Our findings showed optimal R Precision Top 1 at $\sigma=0.2$, best FID at $\sigma=0.1$, and peak diversity at $\sigma=0.4$. While increasing $\sigma$ did enhance MModality, it also resulted in some semantically incorrect motions at higher values. Balancing all factors, we opted for $\sigma=0.2$ in our experimental setup.

\section{Conclusion and Limitation }
In conclusion, our research advances human motion generation with HuTuMotion, an innovative method that utilizes latent diffusion models and few-shot human feedback. Unlike traditional methods, HuTuMotion uniquely adjusts the prior distribution based on human feedback, enhancing data realism. We have also established a feedback mechanism that ensures semantic alignment between motion descriptions and corresponding latent distributions. Importantly, HuTuMotion accommodates personalized and style-aware motion generation, broadening its potential applications. Quantitative and qualitative experiments affirm HuTuMotion's superiority over existing methods. Due to the higher dimensionality of an explicit diffusion model (e.g., MDM) is hard to optimize, the proposed method is limited to latent motion diffusion (i.e. MLD). In additiion, when processing long prompts with numerous action descriptions, MLD tends to miss some actions. Our method to improve this issue is relatively limited.


\section{Acknowledgments}
This work was supported by Key Research and Development projects in Shaanxi Province (No. 2023-YBNY-121), Xi'an Science and Technology Plan Project (No. 22NYYF013), Xianyang Key Project of Research and Development Plan (No. L2022ZDYFSF050), and the National Key Research and Development Program of China (No. 2021YFD1600704).

\bibliography{aaai24}

\begin{thebibliography}{49}
\providecommand{\natexlab}[1]{#1}

\bibitem[{Ahuja and Morency(2019)}]{ahuja2019language2pose}
Ahuja, C.; and Morency, L.-P. 2019.
\newblock Language2pose: Natural language grounded pose forecasting.
\newblock In \emph{2019 International Conference on 3D Vision (3DV)}, 719--728.
  IEEE.

\bibitem[{Bhattacharya et~al.(2021)Bhattacharya, Rewkowski, Banerjee, Guhan,
  Bera, and Manocha}]{bhattacharya2021text2gestures}
Bhattacharya, U.; Rewkowski, N.; Banerjee, A.; Guhan, P.; Bera, A.; and
  Manocha, D. 2021.
\newblock Text2gestures: A transformer-based network for generating emotive
  body gestures for virtual agents.
\newblock In \emph{2021 IEEE Virtual Reality and 3D User Interfaces (VR)},
  1--10. IEEE.

\bibitem[{Chen et~al.(2023)Chen, Jiang, Liu, Huang, Fu, Chen, and
  Yu}]{chen2022mld}
Chen, X.; Jiang, B.; Liu, W.; Huang, Z.; Fu, B.; Chen, T.; and Yu, G. 2023.
\newblock Executing Your Commands via Motion Diffusion in Latent Space.
\newblock In \emph{Proceedings of the IEEE/CVF Conference on Computer Vision
  and Pattern Recognition (CVPR)}, 18000--18010.

\bibitem[{Chen et~al.(2018)Chen, Fu, Zhang, Jiang, Xue, and
  Sigal}]{Chen2018MultiLevelSF}
Chen, Z.; Fu, Y.; Zhang, Y.; Jiang, Y.-G.; Xue, X.; and Sigal, L. 2018.
\newblock Multi-Level Semantic Feature Augmentation for One-Shot Learning.
\newblock \emph{IEEE Transactions on Image Processing}, 28: 4594--4605.

\bibitem[{Cheng et~al.(2023)Cheng, Huang, Ning, and Shan}]{cheng2023bopr}
Cheng, Y.; Huang, S.; Ning, J.; and Shan, Y. 2023.
\newblock BoPR: Body-aware Part Regressor for Human Shape and Pose Estimation.
\newblock \emph{arXiv preprint arXiv:2303.11675}.

\bibitem[{Finn, Abbeel, and Levine(2017)}]{MAML}
Finn, C.; Abbeel, P.; and Levine, S. 2017.
\newblock Model-Agnostic Meta-Learning for Fast Adaptation of Deep Networks.
\newblock In \emph{ICML}.

\bibitem[{Gal et~al.(2023)Gal, Alaluf, Atzmon, Patashnik, Bermano, Chechik, and
  Cohen-Or}]{gal2022image}
Gal, R.; Alaluf, Y.; Atzmon, Y.; Patashnik, O.; Bermano, A.~H.; Chechik, G.;
  and Cohen-Or, D. 2023.
\newblock An image is worth one word: Personalizing text-to-image generation
  using textual inversion.
\newblock \emph{ICLR}.

\bibitem[{Ghosh et~al.(2021)Ghosh, Cheema, Oguz, Theobalt, and
  Slusallek}]{ghosh2021synthesis}
Ghosh, A.; Cheema, N.; Oguz, C.; Theobalt, C.; and Slusallek, P. 2021.
\newblock Synthesis of compositional animations from textual descriptions.
\newblock In \emph{Proceedings of the IEEE/CVF International Conference on
  Computer Vision}, 1396--1406.

\bibitem[{Guo et~al.(2022{\natexlab{a}})Guo, Zou, Zuo, Wang, Ji, Li, and
  Cheng}]{Guo_2022_CVPR_t2m}
Guo, C.; Zou, S.; Zuo, X.; Wang, S.; Ji, W.; Li, X.; and Cheng, L.
  2022{\natexlab{a}}.
\newblock Generating Diverse and Natural 3D Human Motions From Text.
\newblock In \emph{Proceedings of the IEEE/CVF Conference on Computer Vision
  and Pattern Recognition (CVPR)}, 5152--5161.

\bibitem[{Guo et~al.(2022{\natexlab{b}})Guo, Zou, Zuo, Wang, Ji, Li, and
  Cheng}]{Guo_2022_CVPR_humanml3d}
Guo, C.; Zou, S.; Zuo, X.; Wang, S.; Ji, W.; Li, X.; and Cheng, L.
  2022{\natexlab{b}}.
\newblock Generating Diverse and Natural 3D Human Motions From Text.
\newblock In \emph{Proceedings of the IEEE/CVF Conference on Computer Vision
  and Pattern Recognition (CVPR)}, 5152--5161.

\bibitem[{Guo et~al.(2022{\natexlab{c}})Guo, Zuo, Wang, and
  Cheng}]{chuan2022tm2t}
Guo, C.; Zuo, X.; Wang, S.; and Cheng, L. 2022{\natexlab{c}}.
\newblock TM2T: Stochastic and Tokenized Modeling for the Reciprocal Generation
  of 3D Human Motions and Texts.
\newblock In \emph{ECCV}.

\bibitem[{Guo et~al.(2020)Guo, Zuo, Wang, Zou, Sun, Deng, Gong, and
  Cheng}]{guo2020action2motion}
Guo, C.; Zuo, X.; Wang, S.; Zou, S.; Sun, Q.; Deng, A.; Gong, M.; and Cheng, L.
  2020.
\newblock Action2motion: Conditioned generation of 3d human motions.
\newblock In \emph{Proceedings of the 28th ACM International Conference on
  Multimedia}, 2021--2029.

\bibitem[{Ho and Salimans(2021)}]{ho2022classifier}
Ho, J.; and Salimans, T. 2021.
\newblock Classifier-free diffusion guidance.
\newblock \emph{NeurIPS workshop on Deep Generative Models and Downstream
  Applications}.

\bibitem[{Hu et~al.(2022)Hu, Li, St{\"u}hmer, Kim, and
  Hospedales}]{hu2022pushing}
Hu, S.~X.; Li, D.; St{\"u}hmer, J.; Kim, M.; and Hospedales, T.~M. 2022.
\newblock Pushing the Limits of Simple Pipelines for Few-Shot Learning:
  External Data and Fine-Tuning Make a Difference.
\newblock In \emph{CVPR}.

\bibitem[{Kanazawa et~al.(2018)Kanazawa, Black, Jacobs, and
  Malik}]{kanazawa2018end}
Kanazawa, A.; Black, M.~J.; Jacobs, D.~W.; and Malik, J. 2018.
\newblock End-to-end recovery of human shape and pose.
\newblock In \emph{Proceedings of the IEEE conference on computer vision and
  pattern recognition}, 7122--7131.

\bibitem[{Lazarou, Stathaki, and Avrithis(2022)}]{lazarou2022tensor}
Lazarou, M.; Stathaki, T.; and Avrithis, Y. 2022.
\newblock Tensor feature hallucination for few-shot learning.
\newblock In \emph{WACV}.

\bibitem[{Lee et~al.(2019)Lee, Yang, Liu, Wang, Lu, Yang, and
  Kautz}]{lee2019dancing}
Lee, H.-Y.; Yang, X.; Liu, M.-Y.; Wang, T.-C.; Lu, Y.-D.; Yang, M.-H.; and
  Kautz, J. 2019.
\newblock Dancing to music.
\newblock \emph{Advances in neural information processing systems}, 32.

\bibitem[{Lee et~al.(2023)Lee, Liu, Ryu, Watkins, Du, Boutilier, Abbeel,
  Ghavamzadeh, and Gu}]{lee2023aligning}
Lee, K.; Liu, H.; Ryu, M.; Watkins, O.; Du, Y.; Boutilier, C.; Abbeel, P.;
  Ghavamzadeh, M.; and Gu, S.~S. 2023.
\newblock Aligning text-to-image models using human feedback.
\newblock \emph{arXiv preprint arXiv:2302.12192}.

\bibitem[{Li et~al.(2022)Li, Zhao, Zhelun, and Sheng}]{li2022danceformer}
Li, B.; Zhao, Y.; Zhelun, S.; and Sheng, L. 2022.
\newblock Danceformer: Music conditioned 3d dance generation with parametric
  motion transformer.
\newblock In \emph{Proceedings of the AAAI Conference on Artificial
  Intelligence}, volume~36, 1272--1279.

\bibitem[{Ling et~al.(2020)Ling, Zinno, Cheng, and van~de
  Panne}]{motionvae_ling2020character}
Ling, H.~Y.; Zinno, F.; Cheng, G.; and van~de Panne, M. 2020.
\newblock Character Controllers Using Motion VAEs.
\newblock \emph{ACM Trans. Graph.}, 39(4).

\bibitem[{Lu et~al.(2022{\natexlab{a}})Lu, Zhou, Bao, Chen, Li, and
  Zhu}]{lu2022dpm}
Lu, C.; Zhou, Y.; Bao, F.; Chen, J.; Li, C.; and Zhu, J. 2022{\natexlab{a}}.
\newblock Dpm-solver: A fast ode solver for diffusion probabilistic model
  sampling in around 10 steps.
\newblock \emph{Advances in Neural Information Processing Systems}, 35:
  5775--5787.

\bibitem[{Lu et~al.(2022{\natexlab{b}})Lu, Zhou, Bao, Chen, Li, and
  Zhu}]{lu2022dpm++}
Lu, C.; Zhou, Y.; Bao, F.; Chen, J.; Li, C.; and Zhu, J. 2022{\natexlab{b}}.
\newblock Dpm-solver++: Fast solver for guided sampling of diffusion
  probabilistic models.
\newblock \emph{arXiv preprint arXiv:2211.01095}.

\bibitem[{OpenAI(2022)}]{chatgpt}
OpenAI. 2022.
\newblock ChatGPT,https://openai.com/ blog/chatgpt/.

\bibitem[{Oreshkin, López, and Lacoste(2018)}]{TADAM}
Oreshkin, B.~N.; López, P.~R.; and Lacoste, A. 2018.
\newblock TADAM: Task dependent adaptive metric for improved few-shot learning.
\newblock In \emph{NeurIPS}.

\bibitem[{Ouyang et~al.(2022)Ouyang, Wu, Jiang, Almeida, Wainwright, Mishkin,
  Zhang, Agarwal, Slama, Ray et~al.}]{ouyang2022traininginstructgpt}
Ouyang, L.; Wu, J.; Jiang, X.; Almeida, D.; Wainwright, C.; Mishkin, P.; Zhang,
  C.; Agarwal, S.; Slama, K.; Ray, A.; et~al. 2022.
\newblock Training language models to follow instructions with human feedback.
\newblock \emph{Advances in Neural Information Processing Systems}, 35:
  27730--27744.

\bibitem[{Petrovich, Black, and Varol(2021)}]{petrovich21actor}
Petrovich, M.; Black, M.~J.; and Varol, G. 2021.
\newblock Action-Conditioned 3{D} Human Motion Synthesis with Transformer
  {VAE}.
\newblock In \emph{International Conference on Computer Vision (ICCV)}.

\bibitem[{Petrovich, Black, and Varol(2022)}]{petrovich22temos}
Petrovich, M.; Black, M.~J.; and Varol, G. 2022.
\newblock {TEMOS}: Generating diverse human motions from textual descriptions.
\newblock In \emph{European Conference on Computer Vision ({ECCV})}.

\bibitem[{Plappert, Mandery, and Asfour(2016)}]{Plappert2016kit}
Plappert, M.; Mandery, C.; and Asfour, T. 2016.
\newblock The KIT Motion-Language Dataset.
\newblock \emph{Big Data}, 4(4): 236--252.

\bibitem[{Plappert, Mandery, and Asfour(2018)}]{plappert2018learning}
Plappert, M.; Mandery, C.; and Asfour, T. 2018.
\newblock Learning a bidirectional mapping between human whole-body motion and
  natural language using deep recurrent neural networks.
\newblock \emph{Robotics and Autonomous Systems}, 109: 13--26.

\bibitem[{Raab et~al.(2022)Raab, Leibovitch, Li, Aberman, Sorkine-Hornung, and
  Cohen-Or}]{raab2022modi}
Raab, S.; Leibovitch, I.; Li, P.; Aberman, K.; Sorkine-Hornung, O.; and
  Cohen-Or, D. 2022.
\newblock MoDi: Unconditional Motion Synthesis from Diverse Data.
\newblock \emph{arXiv preprint arXiv:2206.08010}.

\bibitem[{Radford et~al.(2021)Radford, Kim, Hallacy, Ramesh, Goh, Agarwal,
  Sastry, Askell, Mishkin, Clark et~al.}]{radford2021learningCLIP}
Radford, A.; Kim, J.~W.; Hallacy, C.; Ramesh, A.; Goh, G.; Agarwal, S.; Sastry,
  G.; Askell, A.; Mishkin, P.; Clark, J.; et~al. 2021.
\newblock Learning transferable visual models from natural language
  supervision.
\newblock In \emph{ICML}.

\bibitem[{Rombach et~al.(2022{\natexlab{a}})Rombach, Blattmann, Lorenz, Esser,
  and Ommer}]{rombach2022high}
Rombach, R.; Blattmann, A.; Lorenz, D.; Esser, P.; and Ommer, B.
  2022{\natexlab{a}}.
\newblock High-resolution image synthesis with latent diffusion models.
\newblock In \emph{Proceedings of the IEEE/CVF Conference on Computer Vision
  and Pattern Recognition}, 10684--10695.

\bibitem[{Rombach et~al.(2022{\natexlab{b}})Rombach, Blattmann, Lorenz, Esser,
  and Ommer}]{stable_diffusion}
Rombach, R.; Blattmann, A.; Lorenz, D.; Esser, P.; and Ommer, B.
  2022{\natexlab{b}}.
\newblock High-Resolution Image Synthesis with Latent Diffusion Models.
\newblock In \emph{Proceedings of the IEEE Conference on Computer Vision and
  Pattern Recognition (CVPR)}.

\bibitem[{Ruiz et~al.(2023)Ruiz, Li, Jampani, Pritch, Rubinstein, and
  Aberman}]{ruiz2022dreambooth}
Ruiz, N.; Li, Y.; Jampani, V.; Pritch, Y.; Rubinstein, M.; and Aberman, K.
  2023.
\newblock Dreambooth: Fine tuning text-to-image diffusion models for
  subject-driven generation.
\newblock \emph{CVPR}.

\bibitem[{Samuel et~al.(2023)Samuel, Ben-Ari, Raviv, Darshan, and
  Chechik}]{samuel2023allseedselect}
Samuel, D.; Ben-Ari, R.; Raviv, S.; Darshan, N.; and Chechik, G. 2023.
\newblock It is all about where you start: Text-to-image generation with seed
  selection.
\newblock \emph{arXiv preprint arXiv:2304.14530}.

\bibitem[{Song, Meng, and Ermon(2020)}]{song2020denoisingddim}
Song, J.; Meng, C.; and Ermon, S. 2020.
\newblock Denoising diffusion implicit models.
\newblock \emph{arXiv preprint arXiv:2010.02502}.

\bibitem[{Stiennon et~al.(2020)Stiennon, Ouyang, Wu, Ziegler, Lowe, Voss,
  Radford, Amodei, and Christiano}]{stiennon2020learningsummarize}
Stiennon, N.; Ouyang, L.; Wu, J.; Ziegler, D.; Lowe, R.; Voss, C.; Radford, A.;
  Amodei, D.; and Christiano, P.~F. 2020.
\newblock Learning to summarize with human feedback.
\newblock \emph{Advances in Neural Information Processing Systems}, 33:
  3008--3021.

\bibitem[{Tang, Rybin, and Chang(2023)}]{tang2023zeroth}
Tang, Z.; Rybin, D.; and Chang, T.-H. 2023.
\newblock Zeroth-Order Optimization Meets Human Feedback: Provable Learning via
  Ranking Oracles.
\newblock \emph{arXiv preprint arXiv:2303.03751}.

\bibitem[{Tevet et~al.(2022)Tevet, Raab, Gordon, Shafir, Bermano, and
  Cohen-Or}]{mdm2022human}
Tevet, G.; Raab, S.; Gordon, B.; Shafir, Y.; Bermano, A.~H.; and Cohen-Or, D.
  2022.
\newblock Human Motion Diffusion Model.
\newblock \emph{arXiv preprint arXiv:2209.14916}.

\bibitem[{Vinyals et~al.(2016)Vinyals, Blundell, Lillicrap, kavukcuoglu, and
  Wierstra}]{MatchNet}
Vinyals, O.; Blundell, C.; Lillicrap, T.; kavukcuoglu, k.; and Wierstra, D.
  2016.
\newblock Matching Networks for One Shot Learning.
\newblock In \emph{NeurIPS}.

\bibitem[{Wu et~al.(2023)Wu, Sun, Zhu, Zhao, and Li}]{wu2023better}
Wu, X.; Sun, K.; Zhu, F.; Zhao, R.; and Li, H. 2023.
\newblock Better Aligning Text-to-Image Models with Human Preference.
\newblock \emph{arXiv preprint arXiv:2303.14420}.

\bibitem[{Xu et~al.(2023)Xu, Liu, Wu, Tong, Li, Ding, Tang, and
  Dong}]{xu2023imagereward}
Xu, J.; Liu, X.; Wu, Y.; Tong, Y.; Li, Q.; Ding, M.; Tang, J.; and Dong, Y.
  2023.
\newblock ImageReward: Learning and Evaluating Human Preferences for
  Text-to-Image Generation.
\newblock \emph{arXiv preprint arXiv:2304.05977}.

\bibitem[{Yang, Wang, and Zhu(2022)}]{yang2022few}
Yang, Z.; Wang, J.; and Zhu, Y. 2022.
\newblock Few-Shot Classification with Contrastive Learning.
\newblock In \emph{ECCV}.

\bibitem[{Ye et~al.(2020)Ye, Hu, Zhan, and Sha}]{Ye_2020_CVPR}
Ye, H.-J.; Hu, H.; Zhan, D.-C.; and Sha, F. 2020.
\newblock Few-Shot Learning via Embedding Adaptation With Set-to-Set Functions.
\newblock In \emph{CVPR}.

\bibitem[{Yu et~al.(2023)Yu, Huang, Fang, Breckon, and Wang}]{yu2023acr}
Yu, Z.; Huang, S.; Fang, C.; Breckon, T.~P.; and Wang, J. 2023.
\newblock ACR: Attention Collaboration-based Regressor for Arbitrary Two-Hand
  Reconstruction.
\newblock In \emph{Proceedings of the IEEE/CVF Conference on Computer Vision
  and Pattern Recognition}, 12955--12964.

\bibitem[{Zhang et~al.(2023{\natexlab{a}})Zhang, Huang, Tu, Chen, Zhan, Yu, and
  Shan}]{zhang2023tapmo}
Zhang, J.; Huang, S.; Tu, Z.; Chen, X.; Zhan, X.; Yu, G.; and Shan, Y.
  2023{\natexlab{a}}.
\newblock TapMo: Shape-aware Motion Generation of Skeleton-free Characters.
\newblock \emph{arXiv preprint arXiv:2310.12678}.

\bibitem[{Zhang et~al.(2022{\natexlab{a}})Zhang, Cai, Pan, Hong, Guo, Yang, and
  Liu}]{zhang2022motiondiffuse}
Zhang, M.; Cai, Z.; Pan, L.; Hong, F.; Guo, X.; Yang, L.; and Liu, Z.
  2022{\natexlab{a}}.
\newblock MotionDiffuse: Text-Driven Human Motion Generation with Diffusion
  Model.
\newblock \emph{arXiv preprint arXiv:2208.15001}.

\bibitem[{Zhang et~al.(2022{\natexlab{b}})Zhang, Zhang, Fang, Gao, Li, Dai,
  Qiao, and Li}]{zhang2022tip}
Zhang, R.; Zhang, W.; Fang, R.; Gao, P.; Li, K.; Dai, J.; Qiao, Y.; and Li, H.
  2022{\natexlab{b}}.
\newblock Tip-Adapter: Training-free Adaption of CLIP for Few-shot
  Classification.
\newblock In \emph{ECCV}.

\bibitem[{Zhang et~al.(2023{\natexlab{b}})Zhang, Yang, Feng, Qin, Chen, Yu,
  Chen, Wang, Savarese, Ermon et~al.}]{zhang2023hive}
Zhang, S.; Yang, X.; Feng, Y.; Qin, C.; Chen, C.-C.; Yu, N.; Chen, Z.; Wang,
  H.; Savarese, S.; Ermon, S.; et~al. 2023{\natexlab{b}}.
\newblock HIVE: Harnessing Human Feedback for Instructional Visual Editing.
\newblock \emph{arXiv preprint arXiv:2303.09618}.

\end{thebibliography}


\begin{figure}[h]
	\centering
	\includegraphics[width=0.99\linewidth]{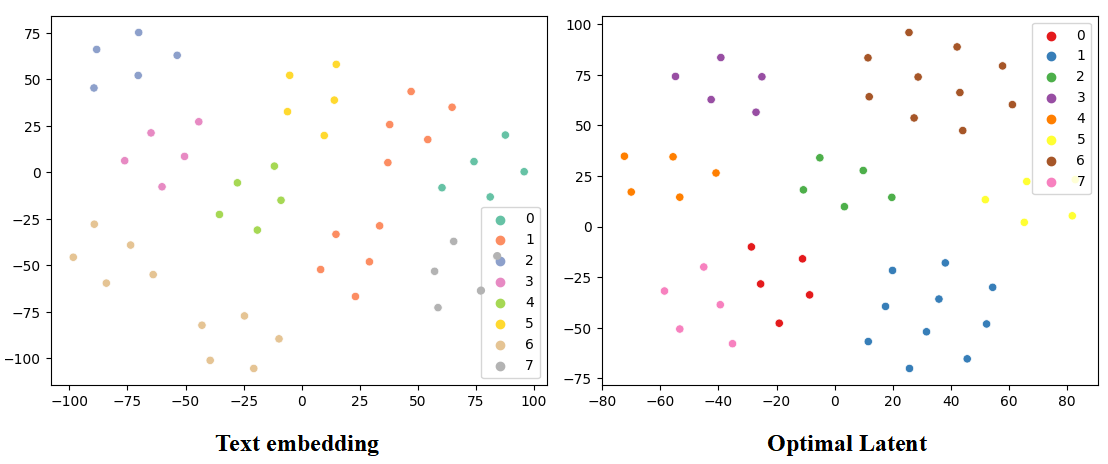}
	\caption{Text embedding (left) and optimal latent distribution (right) clustering by t-SNE. We initially applied spectral clustering to random 50 text embeddings, categorizing them into 8 distinct classes based on cosine similarity matrix. Subsequently, we employed the labels assigned to the text embeddings to visualize the t-SNE results of the 50 optimal latent distributions.}
	\label{fig:observation}
\end{figure}

\section{Observation between Text Embedding and Optimal Latent Distribution}

Figure \ref{fig:observation}  showcases the t-SNE visualization of the optimal latent distribution, with the categorized text embedding labels as a reference. The optimal latent distribution is obtained by latent optimization via human feedback. The visualization results affirm the alignment between the text embedding space and the optimal latent distribution space. This finding further supports our ability to utilize a minimal representative latent distribution to guide the latent diffusion model based on text embedding similarity.

\begin{figure}[h]
	\centering
	\includegraphics[width=0.95\linewidth]{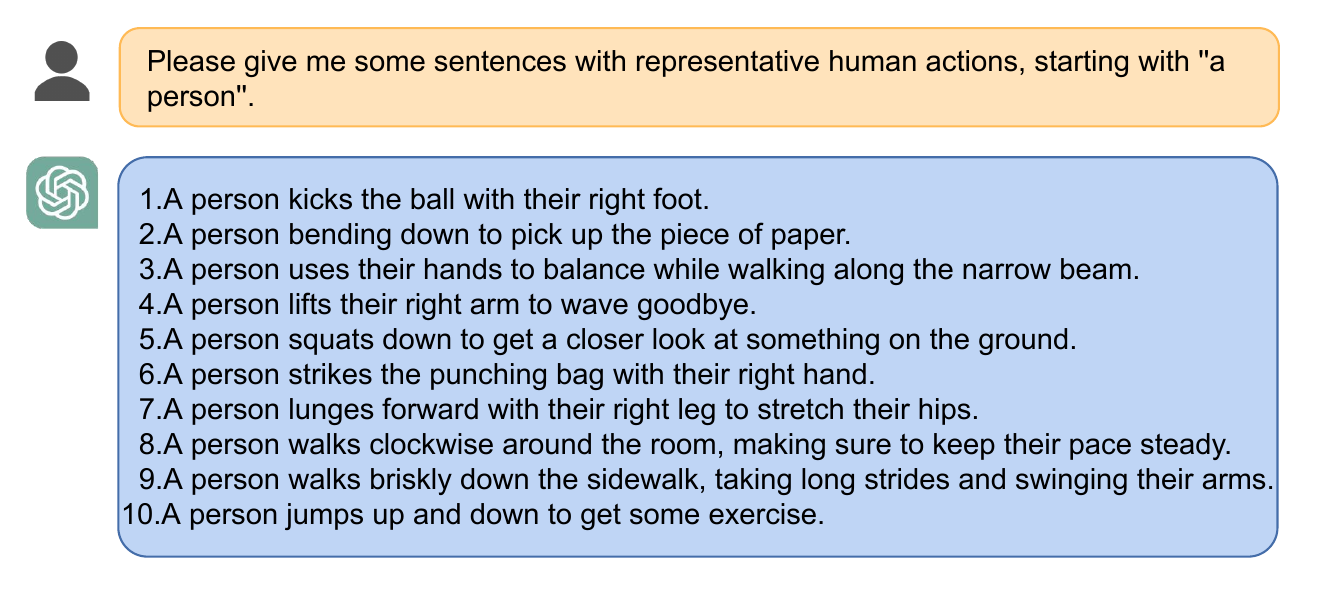}
	\caption{A dialogue with ChatGPT for generating a set of diverse and representative human action sentences.}
	\label{fig:chatgpt}
\end{figure}

\section{Obtaining Representative Texts by Asking ChatGPT}
By harnessing the robust text generation capabilities of large language models (LLMs), we employ ChatGPT \cite{chatgpt} to generate a diverse set of exemplary texts, as depicted in Fig. \ref{fig:chatgpt}. For our implementation, we utilize the initial 5 representative samples. Notably, we observe that these generated sentences exhibit representativeness across the humanML3D and KIT datasets. This crucial aspect allows us to acquire representative latent distributions while optimizing the process using human ranking information. As a result, this significantly enhances the generalization of our semantically guided generation approach.

\section{ Experimental Details for Algorithm 1}

In our implementation of Algorithm 1, we have specified the following parameter values: the number of queries, denoted as $m$, is set to 4; the step size, denoted as $\eta$, is set to 1; and the shrinking rate is set to 0.5. Additionally, we set the smoothing parameters $\mu_1$, $\mu_2$, and $\mu_3$ to the values of 0.8, 0.4, and 0.1, respectively.

To determine the weighted $\Xc_1$ using the ranking information $\Ibb_1$, we formulate the calculation as follows:

\begin{equation}
z^{*} = \sum_{i=1}^{m} \text{softmax}(\Ibb_1^i) \cdot \Xc_1^i,
\end{equation}

here, the ranking information $\Ibb_1$ takes on values ranging from 1 to $k$, and we compute the softmax function of each $\Ibb_1^i$ before multiplying it by the corresponding $\Xc_1^i$ term.

\section{ User Interface on Algorithm 1}

\begin{figure}[t]
	\centering
	\includegraphics[width=0.99\linewidth]{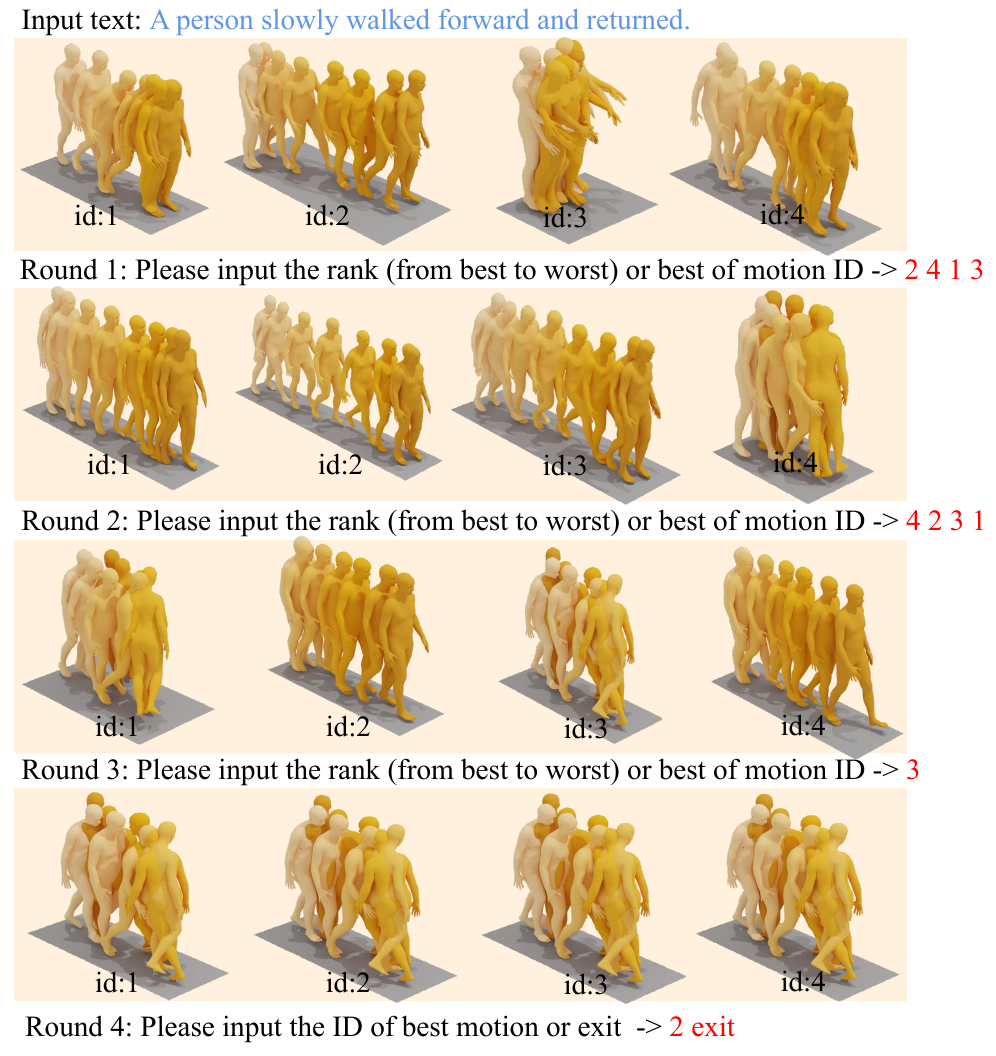}
	\caption{The User Interface of Algorithm 1.}
	\label{fig:ui}
\end{figure}

Figure \ref{fig:ui} presents the corresponding user interface (UI) designed for human ranking feedback in Algorithm 1. We optimize the input text prompt ``A person slowly walked forward and returned." using 4 rounds, and 4 motions are presented to the users at each round.  When the user receives the instruction ``Please input the rank (from best to worst) or best of motion ID," it indicates that the algorithm is in the first stage, which focuses on optimizing the latent distribution by computing gradients based on ranking information. Subsequently, when the user inputs the best motion ID, it signifies that the user has entered the second stage for fine-tuning the latent distribution. The user is prompted with the instruction ``Please input the ID of the best motion," and they only need to select the best motion from the provided motions. This interface facilitates seamless and intuitive communication between the user and the algorithm. Ultimately, we obtain the optimized motion and paired data consisting of text embeddings and optimal latent distributions. It should be noted that the generated motion diversity is influenced by our smoothing parameters which alter the scale of $z$. The larger the smoothing parameters, the greater the diversity. Throughout the representative distribution optimization process, we gradually decrease the smoothing parameters according to human feedback to achieve convergence.

\begin{figure}[h]
	\centering
	\includegraphics[width=0.99\linewidth]{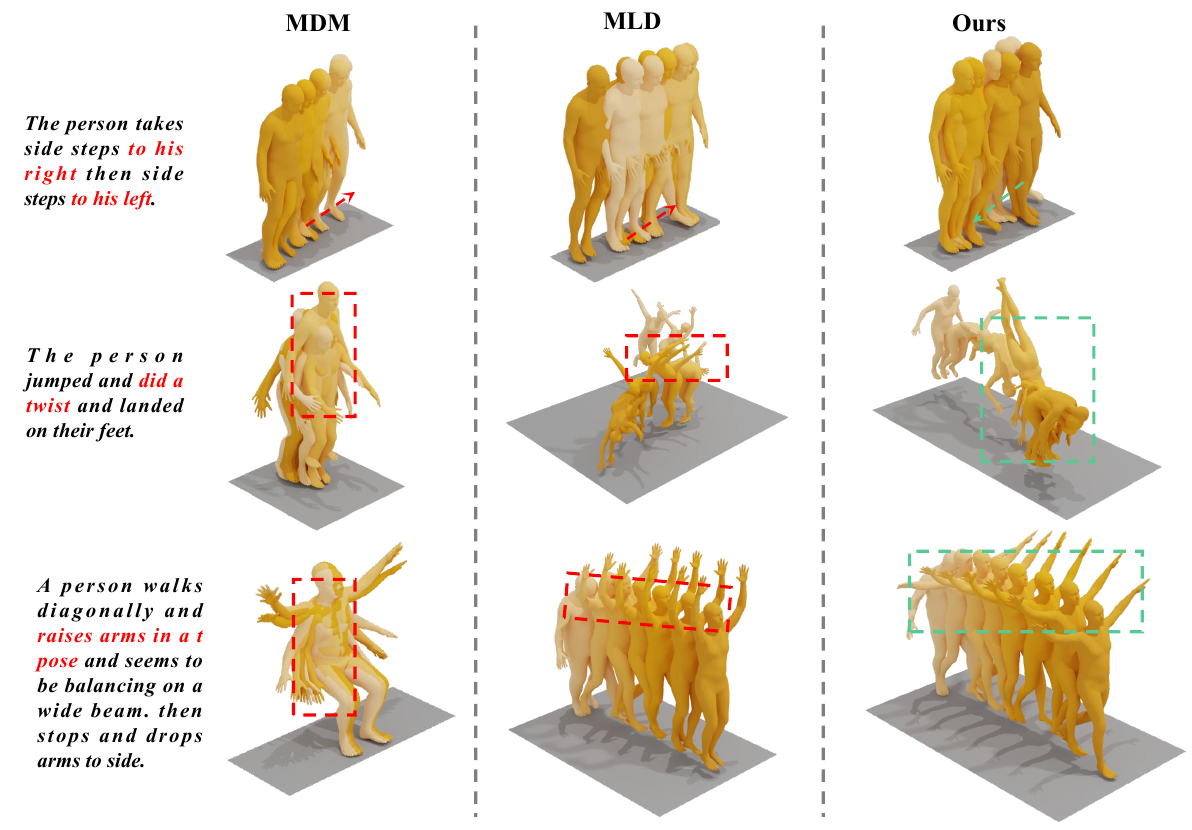}
	\caption{More visualization comparison results. The darker colors indicate the later frame in time.}
	\label{fig:visres2}
\end{figure}

\begin{figure}[h]
	\centering
	\includegraphics[width=0.99\linewidth]{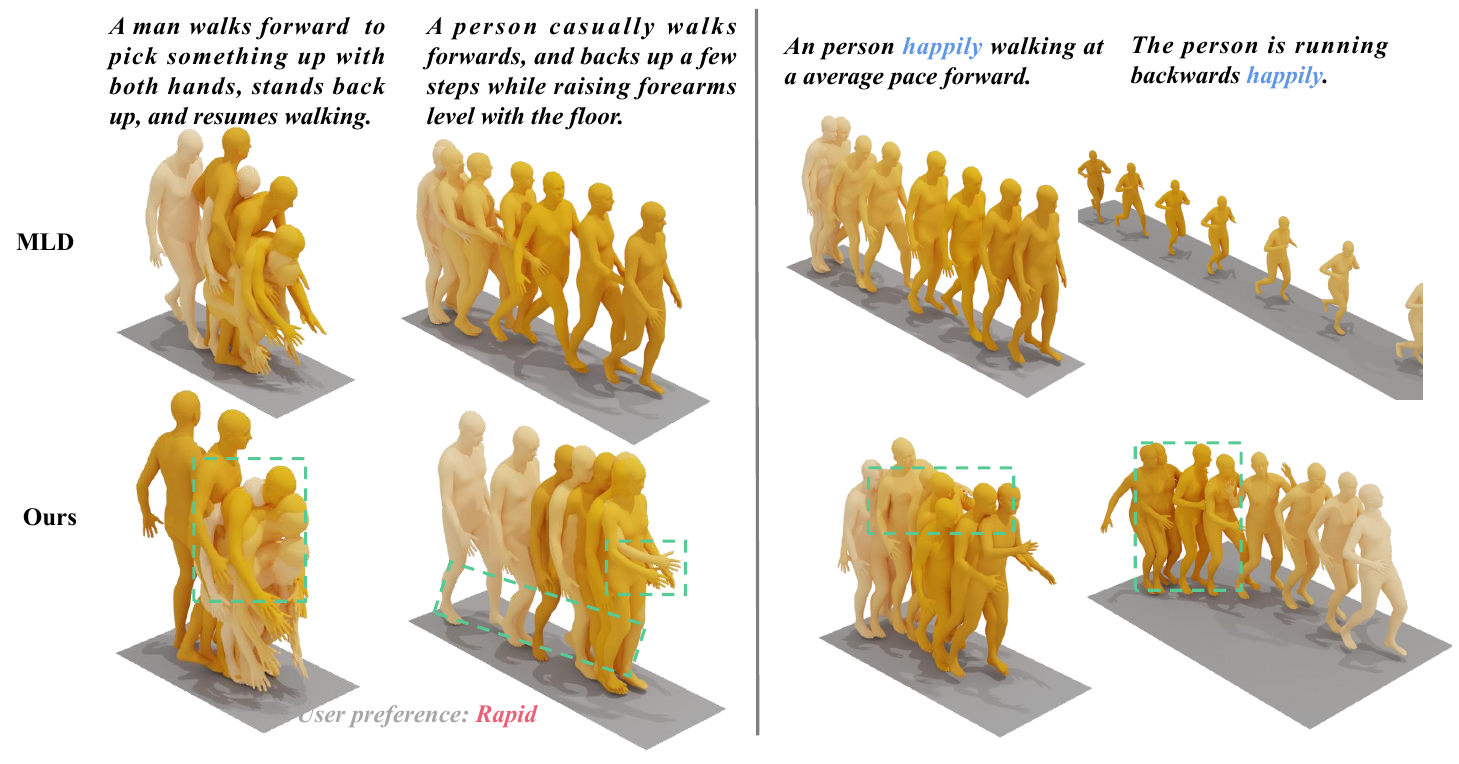}
	\caption{Personalized (left) and style-aware motion generation (right). The darker colors indicate the later frame in time. The darker colors indicate the later frame in time.}
	\label{fig:application2}
\end{figure}

\section{More Results}
Figure \ref{fig:application2} illustrates the generation of personalized and stylized motion, incorporating the user's preference for ``rapid" motion and the desired style of ``happily." Additionally, Figure \ref{fig:visres2} presents additional comparison results based on the HumanML3D test set. These results showcase the superior capability of our method in producing motions that are both natural and semantically accurate, surpassing the performance of MDM \cite{mdm2022human} and MLD \cite{chen2022mld}.

For personalization, using the same text input, MLD exhibits normal motions, while we can generate rapid actions that align with human preferences. In addition, for the given text "a person casually walks forwards, and backs up a few steps while raising both forearms level with the floor", we also exhabit the sequence of ``backs up a few steps," while MLD fails to geenerate.  For text-guided stylization, we observed that MLD completely fails to generate stylized motion. For the given text ``the person is running backwards happily", MLD does not generate a motion sequence with the style of ``happily" while we clearly displaying a happly style.

\end{document}